\documentclass{article}

\PassOptionsToPackage{numbers, compress}{natbib}

\usepackage[preprint]{neurips_2023}

\usepackage[utf8]{inputenc} 
\usepackage[T1]{fontenc}    
\usepackage{hyperref}       
\usepackage{url}            
\usepackage{booktabs}       
\usepackage{amsfonts}       
\usepackage{nicefrac}       
\usepackage{microtype}      
\usepackage{xcolor}         

\usepackage{tabularx}
\usepackage{booktabs}
\usepackage{threeparttable}
\usepackage{subfigure}
\usepackage{todonotes}
\usepackage{amsmath}
\usepackage{mathtools}
\usepackage{booktabs}
\usepackage{amsfonts}
\usepackage{pifont}
\usepackage{wasysym}
\usepackage{multirow}
\usepackage{chemfig}

\title{Multi-scale Transformer Pyramid Networks for Multivariate Time Series Forecasting}

\author{%
  Yifan Zhang\thanks{Corresponding Author}, Sergiu M. Dascalu, Frederick C. Harris, Jr.\\
  Department of Computer Science and Engineering,
  University of Nevada, Reno\\
  \texttt{yfzhang@nevada.unr.edu},\\ \texttt{\{dascalus,Fred.Harris\}@cse.unr.edu} 
  \And
  Rui Wu \\
  Department of Computer Science, 
  East Carolina University\\
  \texttt{WUR18@ecu.edu}
}

\begin{document}

\maketitle

\begin{abstract}
Multivariate Time Series (MTS) forecasting involves modeling temporal dependencies within historical records. Transformers have demonstrated remarkable performance in MTS forecasting due to their capability to capture long-term dependencies. However, prior work has been confined to modeling temporal dependencies at either a fixed scale or multiple scales that exponentially increase (most with base 2). This limitation hinders their effectiveness in capturing diverse seasonalities, such as hourly and daily patterns. In this paper, we introduce a dimension invariant embedding technique that captures short-term temporal dependencies and projects MTS data into a higher-dimensional space, while preserving the dimensions of time steps and variables in MTS data. Furthermore, we present a novel Multi-scale Transformer Pyramid Network (MTPNet), specifically designed to effectively capture temporal dependencies at multiple unconstrained scales. The predictions are inferred from multi-scale latent representations obtained from transformers at various scales. Extensive experiments on nine benchmark datasets demonstrate that the proposed MTPNet outperforms recent state-of-the-art methods.
\end{abstract}

\section{Introduction}

Multivariate time series (MTS) data, which captures multiple variables over time, is of critical importance in various fields including finance, climate, and energy.
MTS forecasting, a critical machine learning task, aims to predict the future values of multiple variables based on their historical records. 
The MTS data inherently demonstrates low semantic characteristics, necessitating its analysis as a collection of multiple values. For instance, this involves analyzing values of multiple variables at a single time step or values of multiple time steps for a single variable. This approach facilitates the extraction of two types of information from the MTS data: correlations among variables (spatial dependencies) and correlations across time steps (temporal dependencies). 

In recent years, machine learning models, notably transformers~\cite{wen2023transformers_in_TS}, have significantly advanced the exploration of MTS forecasting problems.
The pioneering work by~\cite{NIPS2019transformer_ts_forecasting} introduced transformers to MTS forecasting, highlighting their potential in adeptly capturing temporal dependencies.
Consequently, numerous transformer-based methods have been introduced for the task of MTS forecasting~\cite{haoyietal-informer-2021, wu2021autoformer, Yuqietal-2023-PatchTST}.
A common technique for extracting spatial dependencies utilizes a linear layer to project the MTS data into a higher-dimensional space along the spatial dimension. As a result, the values of variables at a single time step are represented as a vector in this higher-dimensional space.
As for temporal dependencies, their scales are critical in achieving accurate MTS forecasting. However, most existing methods are confined to capturing temporal dependencies solely at a single scale.
For example, Informer~\cite{haoyietal-informer-2021} aims to model temporal dependency between individual time steps in the time series sequence. Autoformer~\cite{wu2021autoformer} proposes an auto-correlation mechanism to capture temporal dependencies among sub-series. PatchTST~\cite{Yuqietal-2023-PatchTST} and Crossformer~\cite{zhang2023crossformer} introduce a patch procedure that divides each series within the MTS data into patches of a specific length, enabling the use of a canonical transformer to model temporal dependencies at the sub-series level.

Few methods have been proposed with the objective of modeling multi-scale temporal dependencies.  
SCINet~\cite{liu2022SCINet} leverages an SCI-Block, which employs downsampling techniques to divide the input sequence into two sub-sequences. This division enables the extraction of distinctive temporal relations at a coarser scale by utilizing two convolution neural network (CNN) kernels. By arranging multiple SCI-Blocks in a hierarchical tree structure across multiple levels, SCINet models temporal relations at various scales.
MICN~\cite{micnICLR2023} also incorporates downsampling techniques, involving the reduction of the original MTS data's resolution. Then, a multi-scale isometric convolution layer, comprising multiple branches of the local-global module, processes downsampled MTS data of varying scale sizes.
Pyraformer~\cite{liu2022pyraformer} introduces a pyramidal graph for modeling MTS data at different resolutions. This pyramidal graph embodies a tree structure in which each parent node has several child nodes. The parent nodes summarize the sub-series of all child nodes. Thus the scale increases exponentially at the base of the number of child nodes.
Crossformer~\cite{zhang2023crossformer} comprises three transformer encoder-decoder pairs that aim to capture temporal dependencies within MTS data at three scales. Crossformer merges the latent representations from the lower level as input for the next level, leading to a doubling of the scale as the levels progress.

However, all those methods suffer from a limitation wherein their multiple scales increase exponentially (most with base 2). Consequently, they may fail to capture certain scales of temporal dependencies in MTS data that are crucial for accurate forecasting tasks. 
This limitation underscores the importance of developing more flexible and adaptable approaches capable of effectively modeling temporal dependencies across a wider range of arbitrary scales within the MTS data.
To address the aforementioned limitations, we propose Multi-scale Transformer Pyramid Networks (MTPNet) that effectively model temporal dependencies at multiple unconstrained scales. 
The contributions of our work are summarized as follows:
\begin{itemize}
    \item We propose a dimension invariant (DI) embedding mechanism that captures short-term temporal dependencies and projects the MTS data into a high-dimensional space. Notably, this DI embedding technique preserves both the spatial and temporal dimensions of the MTS data.
    \item We propose a multi-scale transformer-based pyramid that effectively models temporal dependencies across multiple unconstrained scales, thereby offering the versatility to capture temporal patterns at various resolutions.
    \item We evaluate the proposed MTPNet using nine real-world datasets, and the experimental results demonstrate its superior performance compared to recent state-of-the-art methods.
\end{itemize}

\section{Related works}
\subsection{MTS forecasting}
The primary objective of the MTS forecasting task is to establish an accurate inference between historical observations $X\in \mathbb{R}^{I\times D}$ of $D$ variables within a look-back window of $\textit{I}$ time steps and future $H$ time steps' values of $X_{pred} \in \mathbb{R}^{H\times D}$. 
Traditional statistical methods like ARIMA~\cite{box1968ARIMA} and exponential smoothing~\cite{hyndman2008exponential} are confined to univariate time series forecasting and their performance diminishes as the prediction length $H$ increases. Advances in deep learning have greatly enhanced the development of MTS forecasting. LSTNet~\cite{lai2018lstnet} and TPA-LSTM~\cite{shih2019tpalstm} combine CNN and RNN to capture short-term and long-term temporal dependencies. MTGNN~\cite{wu2020mtgnn} introduces a graph neural network framework explicitly designed to model spatial dependencies among variables in MTS data.
While these methods are based on various neural network architectures, their common objective is to discover forecasting inferences through iterative weight adjustments that minimize the discrepancies between the forecasts and the ground truth.

\subsection{Transformers}
Transformers were first developed for natural language processing~\cite{Vaswani_transformer} and soon achieved great success in computer vision~\cite{dosovitskiy2020vit, wang2021PVT} and MTS forecasting~\cite{NIPS2019transformer_ts_forecasting, haoyietal-informer-2021, wu2021autoformer, Yuqietal-2023-PatchTST, zhang2023crossformer}. The canonical transformer architecture includes a self-attention mechanism and a feed-forward network. To enhance the training process, it employs residual connections~\cite{He_2016_CVPR_resnet} and layer normalization~\cite{ba2016layernorm}. 

Early studies that apply transformers in MTS forecasting focused on modeling temporal dependencies at individual time step resolution. These approaches encountered quadratic time complexity issues, which impose limitations
on the input length—a crucial factor for MTS forecasting.
Informer~\cite{haoyietal-informer-2021} proposed ProbSparse self-attention reduced time complexity to $\mathcal{O}(n\log{}n)$ by only calculating a subset of queries. 
Fedformer~\cite{zhou2022fedformer} enhanced transformers with Fourier transforms and Wavelet transforms and achieved linear computational complexity and memory cost.
Several recent studies~\cite{Yuqietal-2023-PatchTST, zhang2023crossformer} have adopted the patch mechanism introduced in ViT~\cite{dosovitskiy2020vit} to partition the MTS data into patches, thus facilitating the transformer's efficacy in managing extended input sequences and capturing temporal dependencies at the sub-series level.

Despite the successful application of transformers in MTS forecasting, existing methods have limitations in capturing temporal dependencies at various constrained scales. This can hinder their ability to effectively capture seasonality patterns at arbitrary scales.

\section{Method}
\subsection{Decomposition}
We decompose the MTS data into seasonal and trend-cyclical components following ~\cite{wu2021autoformer, zhou2022fedformer, Zeng2022Dlinear}.
Given MTS input $X \in \mathbb{R}^{I\times D}$, the decomposition procedure is as follows:
\begin{equation}\label{eq:decomp}
\begin{split}
\mathbf{X}_{t}&=\operatorname{mean}\left(\sum_{i=1}^{n}\operatorname{AvgPool}\left(\operatorname{Padding}\left({\mathbf{X}}\right)\right)_{i}\right)
 \\
\mathbf{X}_{s}&= \mathbf{X} - \mathbf{X}_{t}
\end{split}
\end{equation}
Where $\mathbf{X}_{s}\in \mathbb{R}^{I\times D}$ and  $\mathbf{X}_{t}\in \mathbb{R}^{I\times D}$ are seasonal and trend-cyclical components, respectively. 

\begin{figure}[!htbp]
    \centering
    \includegraphics[width=0.65\linewidth]{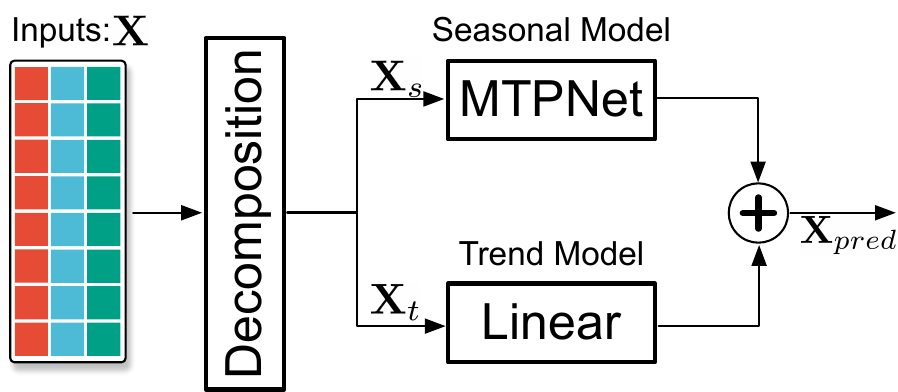}
    \caption{Illustration of the overall framework: Decomposition of MTS data into seasonal and trend-cyclical components, employing Multi-scale Transformer Pyramid Networks (MTPNet) as the seasonal model and a linear layer as the trend model. The seasonal and trend predictions are summed to obtain the final predictions.}
    \label{fig:overall_workflow}
\end{figure}
Figure~\ref{fig:overall_workflow} illustrates our proposed framework, incorporating seasonal and trend models to learn and forecast the seasonal and trend-cyclical components, respectively. The MTPNet functions as the seasonal model, while a simple linear layer is employed as the trend model to infer predictions directly from historical records. In scenarios where the MTS data lacks distinct seasonality and trend, we use MTPNet as the trend model to effectively learn intricate trend-cyclical components.
Finally, the predictions from both the seasonal and trend models are summed elementwise to derive the final MTS predictions. 

\subsection{Transformer Feature Pyramid}
To address the limitations of a transformer that captures temporal dependencies solely at a single scale, we propose a multi-scale transformer pyramid network, as depicted in Figure~\ref{fig:MTPNet}. The primary objective of the MTPNet is to capture temporal dependencies across diverse unconstrained scales, ranging from fine to coarse resolutions. Notably, the total number of levels, denoted by $K$, is not fixed but depends on the array of available patch sizes, where $k = 1, \cdots, K$. This hierarchical architecture empowers MTPNet to model multi-scale representations of the complex temporal dependencies within the input sequence.

As illustrated in Figure~\ref{fig:MTPNet}, transformers at all levels take the MTS sequence as input, which is referred to as all-scale inputs. Note that the decoder inputs are omitted in Figure~\ref{fig:MTPNet} for brevity and details are discussed later. The DI embedding components are distinctive at each level, as they partition input MTS data into patches of unconstrained lengths of $p_k\in\{p_1, \cdots, p_K\}$. Consequently, the multi-level transformers focus on capturing the temporal dependencies at scales from fine to coarse. 

The inter-scale connections facilitate information flow between transformers at different levels within the pyramid architecture. Encoders and decoders are symmetrically structured, with encoders adopting a bottom-up approach and decoders following a top-down pattern. This design allows encoders to progressively learn latent representations from fine to coarse scales, while decoders generate fine-scale representations guided by coarse-scale levels. This yields $K$ latent representations from the feature pyramid. Finally, a 1-layer CNN generates predictions from the concatenated $K$ latent representations.

\subsection{Dimension Invariant Embedding}
This section introduces the DI embedding technique and emphasizes the significance of maintaining both spatial and temporal dimensions intact.
Due to the inherent lack of semantic information in MTS data compared to words or images, transformer-based methods for MTS forecasting commonly group values either along the spatial dimension (variables) or the temporal dimension (time steps) for further analysis.
\begin{figure}[!htbp]
    \centering
    \includegraphics[width=1\linewidth]{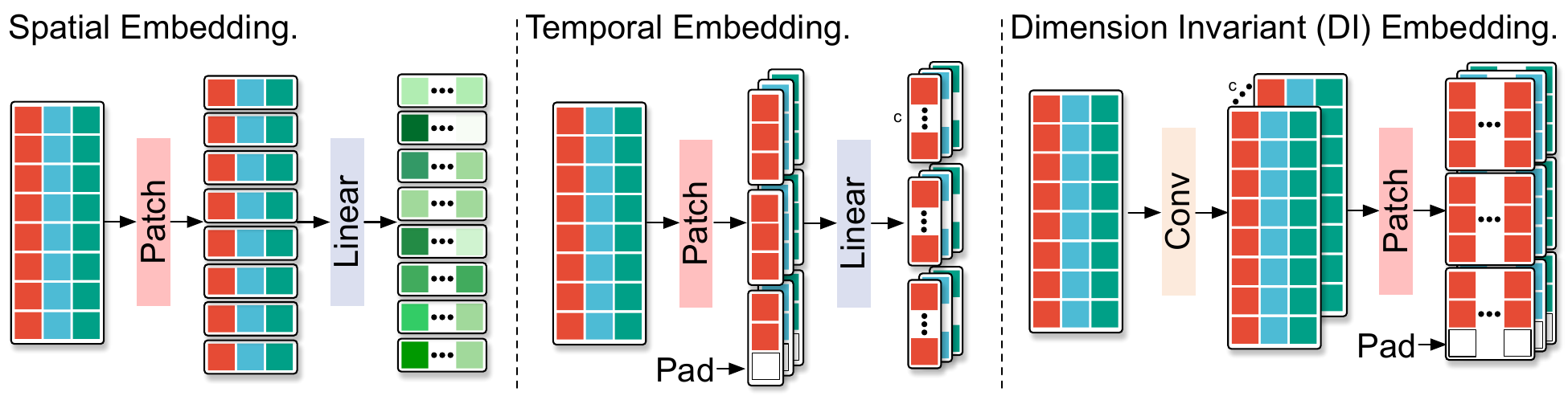}
    \caption{Illustration of spatial, temporal, and dimension invariant embedding techniques.}
    \label{fig:value_embedding}
\end{figure}

Figure~\ref{fig:value_embedding} shows the workflow of spatial, temporal, and DI embedding techniques.
The spatial embedding~\cite{haoyietal-informer-2021} employs a linear layer to project the values of all variables at a single time step into an alternately dimensional space (e.g., 64, 128) while preserving the temporal dimension invariant.
The temporal embedding~\cite{zhang2023crossformer, Yuqietal-2023-PatchTST} preserves the spatial dimension's invariance while employing a linear layer to embed values of a variable at multiple time steps into a higher-dimensional space. 
Both embedding techniques break one dimension of the MTS data: spatial embedding mixes spatial information, while temporal embedding restricts the temporal scale.

To avoid these disadvantages, we introduce the DI embedding technique which utilizes a 1-layer CNN with a kernel size of $3\times1$ to embed the MTS data into feature maps while preserving both spatial and temporal dimensions invariant as follows:
\begin{equation}\label{eq:DI_embed}
\mathbf{X}_{emb}=\operatorname{Conv}\left({\mathbf{X}}_{input}\right)
\end{equation}
where the $\mathbf{X}_{input} \in \mathbb{R}^{1\times I\times D}$ is either $\mathbf{X}_{s}$ (seasonal) or $\mathbf{X}_{t}$ (trend), $\mathbf{X}_{emb} \in \mathbb{R}^{c\times I\times D}$ is the embedded feature maps. The $\operatorname{Conv}$ also captures short-term temporal dependencies.

The DI embedding then applies the $\operatorname{Patch}$ procedure to the embedded inputs, generating patched inputs at scale $p$ as follows:
\begin{equation}\label{eq:patch}
\mathbf{X}_{emb}=\operatorname{Patch}\left(\mathbf{X}_{emb}, \mathbf{X}_{\mathbf{0}}, p\right) \\
\end{equation}
where $\mathbf{X}_{\mathbf{0}}$ is zero-padding if the length of the time series is not divisible by the patch size $p$. The $\operatorname{Patch}$ procedure divides the time series into $N = \lceil I/p \rceil$ non-overlapping patches of size $p$, yielding $\mathbf{X}_{emb} \in \mathbb{R}^{c \times N \times p \times D}$.

\subsection{Transformer Encoder and Decoder}\label{sec:enc_dec}
\begin{figure*}[htbp]
    \centering
    \includegraphics[width=1\linewidth]{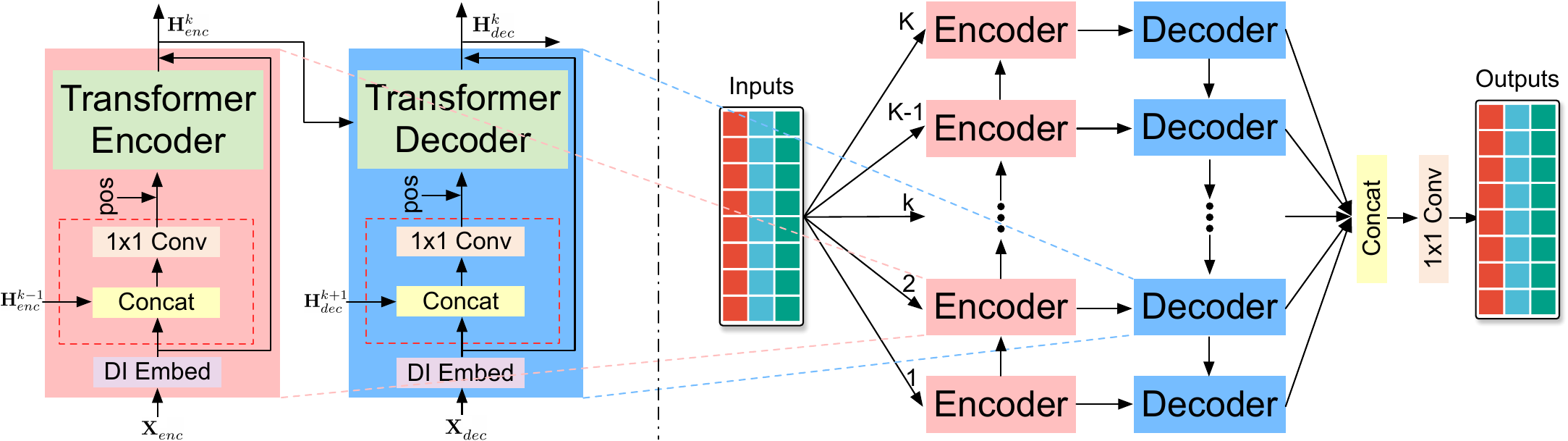}
    \caption{\textbf{Left:} The workflow of a single-level transformer-based encoder-decoder pair. \textbf{Right:} Illustration of the proposed multi-scale transformer pyramid network (MTPNet).}
    \label{fig:MTPNet}
\end{figure*}
The MTPNet comprises multiple transformer encoder-decoder pairs, designed to learn temporal dependencies at multiple unconstrained scales. Figure~\ref{fig:MTPNet} illustrates the detailed computation procedure of one level of the MTPNet. We take the $k$-th level of MTPNet as an illustrative example to elaborate on the detailed computation process of the transformer encoder-decoder pair.

\subsubsection{Encoder:} 
The input $\mathbf{X}_{enc}\in \mathbb{R}^{1\times I\times D}$ are the same for the encoder at any level in the MTPNet. The DI embedding at $k$-th level takes input MTS data and patch size $p_k$ as follows:
\begin{equation}\label{eq:enc_di_embed}
\mathbf{X}_{di}^{k}=\operatorname{DI}\left(\mathbf{X}_{enc}, p_k\right)
\end{equation}
where $\mathbf{X}_{di}^{k}\in \mathbb{R}^{c \times N_{enc}^k \times p_k \times D}$ represents the embeded and patched $k$-th level encoder's input.
Then, the inter-scale connections concatenate and fuse input embedding $\mathbf{X}_{di}^{k}$ with lower level encoder's output $\mathbf{H}_{enc}^{k-1}$ as follows:
\begin{equation}\label{eq:inter_connection}
\mathbf{X}_{emb}^{k}=
\begin{cases}
\mathbf{X}_{di}^{k} & \text{if } k = 1,\\
\operatorname{Conv}\left(\operatorname{Concat}\left(\mathbf{X}_{di}^{k}, \mathbf{H}_{enc}^{k-1}\right)\right) & \text{if } k > 1.
\end{cases}
\end{equation}
where $\operatorname{Concat}$ denote concatenate process along the feature map dimension and $\operatorname{Conv}$ represents 1-layer CNN with a kernel size of $1\times 1$ to fuse and reduce the feature map dimension of concatenated embeddings from $2c$ to $c$.
To incorporate positional information, we add a learnable position embedding $\mathbf{W}_{pos}^{k}$ (denoted pos) to input embedding as follows:
\begin{equation}\label{eq:addpos}
\mathbf{X}_{emb}^{k}=\mathbf{X}_{emb}^{k} + \mathbf{W}_{pos}^{k}
\end{equation}
This step is critical because the inherent nature of the Transformer architecture is order-agnostic. Therefore, positional information needs to be incorporated to capture the temporal dependencies within the input sequence. The input embedding $\mathbf{X}_{emb}^{k}$ is split into univariate embeddings. Therefore, we obtain the transformer encoder's input $\mathbf{X}_{emb}^{k, d}\in \mathbb{R}^{c \times N_{enc}^k \times p_k \times 1}$ which represents the $d$-th variable.

We employ the canonical transformer encoder~\cite{Vaswani_transformer}, utilizing the scaled dot-product attention mechanism as follows:
\begin{equation}\label{eq:attention_QKV}
\begin{split}
Q, K, V &=\operatorname{Linear}\left(\mathbf{X}_{emb}^{k, d}\right) \\
\operatorname{Attention}\left(Q, K, V\right) &= \operatorname{Softmax}\left(QK^T/\sqrt{d_k}\right)V
\end{split}
\end{equation}
where the $Q$, $K$, and $V$ are query, keys, and values embedded from the input sequence of $d$-th series of $\mathbf{X}_{p}^{k, d}\in \mathbb{R}^{c \times N_{enc}^k \times p_k}$. Note that we flatten the feature map and patch size dimension of $\mathbf{X}_{p}^{k, d}$ so that $\mathbf{X}_{p}^{k, d}\in \mathbb{R}^{N_{enc}^k \times (c\times p_k)}$ represents the latent representations of patches of size $p_k$. We also utilize the canonical multi-head attention as follows:
\begin{equation}\label{eq:multihead_attn}
\begin{split}
Q_h, K_h, V_h&=\operatorname{Linear}\left(Q, K, V\right)_h \\
\mathbf{H}_{h}^{k, d}&=\operatorname{Attention}\left(Q_h, K_h, V_h\right) \\
\mathbf{H}_{enc}^{k, d} &= \operatorname{Linear}\left(\operatorname{Concat}\left(\mathbf{H}_{1}^{k}, \cdots, \mathbf{H}_{h}^{k}, \cdots\right)\right)
\end{split}
\end{equation}
where the $\mathbf{H}_{enc}^{k, d}\in \mathbb{R}^{c \times N_{enc}^k \times p_k}$ is the output of transformer encoder at level $k$ for $d$-th series, and the subscript $h$ indicates the $h$-th head of multi-head attention. 
By applying the encoder to all $D$ series in the MTS data, we obtained the $k$-th level's encoder output $\mathbf{H}_{enc}^{k}\in \mathbb{R}^{c \times N_{enc}^k \times p_k\times D}$.
The transformer encoder also includes normalization layers, a feed-forward network, and residual connections, details are available in ~\cite{Vaswani_transformer}. 

The last step of the encoder is skip-connection as follows:
\begin{equation}\label{eq:skip_connection}
\mathbf{H}_{enc}^{k}=\mathbf{H}_{enc}^{k} + \mathbf{X}_{di}^{k}
\end{equation}
where $\mathbf{H}_{enc}^{k}\in \mathbb{R}^{c \times N_{enc}^k \times p_k \times D}$ is the output of $k$-th level encoder.

\subsubsection{Decoder:} 
The decoder's input $X_{dec} \in \mathbb{R}^{(L+H) \times D}$ is the concatenation of the historical records ($L$ time steps) and zero-padding ($H$ future time steps). Similar to the encoder's workflow presented in Equations~\ref{eq:enc_di_embed}, ~\ref{eq:inter_connection}, and~\ref{eq:addpos}, the decoder's input goes through DI embedding and inter-scale connections to obtain a patched embedding for the transformer decoder. It is worth mentioning that the decoder's inter-scale connections follow a top-down order, thus the decoder's output latent representations flow from coarse-scale to fine-scale. Then a learnable position embedding (denoted pos) is added to the input embedding.
Consequently, we obtain the transformer decoder's input, denoted 
$\mathbf{X}_{dec}^{k}\in \mathbb{R}^{c \times N_{dec}^k\times p_k\times D}$. Note that the decoder's number of patches is $N_{dec}^k = \lceil (L+H)/p_k \rceil$.

We also employ the canonical transformer decoder~\cite{Vaswani_transformer}, utilizing the scaled dot-product attention and multi-head attention presented in Equations~\ref{eq:attention_QKV} and~\ref{eq:multihead_attn}. The decoder also includes normalization layers, a feed-forward network, and residual connections, as described in ~\cite{Vaswani_transformer}. The last step of the decoder is also a skip connection as presented in Equation~\ref{eq:skip_connection}.

The output of the decoder at $k$-th level is $\mathbf{H}_{dec}^{k}\in \mathbb{R}^{c \times N_{dec}^k \times p_k\times D}$. To conduct the MTS forecasting task, we only need the latent representations of future time steps. Therefore, we only need the last $N_{pred}^k = \lceil H/p_k \rceil$ patches that presenting future time steps, denoted $\mathbf{H}_{dec}^{k}\in \mathbb{R}^{c \times N_{pred}^k \times p_k\times D}$. At each level, the latent representations represent the predicted value using a vector of length $c$, containing temporal dependencies at a scale of patch size $p_k$.

\subsection{Multi-scale Prediction}
Lastly, we concatenate latent representations of all $K$ levels and generate predictions using a $\operatorname{Conv}$ layer as follows:
\begin{equation}\label{eq:generate_preds}
\begin{split}
\mathbf{H}&=\operatorname{Concat}\left(\mathbf{H}_{dec}^{1}, \cdots,  \mathbf{H}_{dec}^{K}\right) \\
\mathbf{X}_{pred} &= \operatorname{Conv}\left(\mathbf{H}\right)
\end{split}
\end{equation}
To concatenate latent representations at different scales, we apply an inverse patch operation to reassemble patches into the complete sequence. Subsequently, all $K$ latent representations are concatenated, resulting in $\mathbf{H} \in \mathbb{R}^{(K\times c) \times H \times D}$. Each predicting future value is represented as a vector of length $K\times c$, which captures the temporal dependencies at scales ranging from $p_1$ to $p_K$. The $\operatorname{Conv}$ layer project each vector of length $K\times c$ into predicting value, generating the predictions $\mathbf{X}_{pred}\in \mathbb{R}^{H\times D}$. 
The process of inferring predictions from multi-scale latent representations enables the effective utilization of temporal dependencies across a wide range of unconstrained scales, ultimately enhancing forecasting accuracy.

\section{Experiments}

\subsection{Experimental Settings}

\subsubsection{Data:}
In our experiments, we utilized nine publicly available benchmark datasets~\cite{wu2021autoformer}, namely ETTh1, ETTh2, ETTm1, ETTm2, Weather, Traffic, Electricity, Exchange-Rate, and ILI. 
The datasets consist of variables with diverse characteristics, and they are sampled at different frequencies, ranging from every 10 minutes to every 1 week, resulting in varying temporal dependencies among them.
We partition each dataset into training, validation, and test sub-sets with a ratio of 0.6:0.2:0.2 for four ETT datasets and 0.7:0.1:0.2 for the five remaining datasets.

\subsubsection{Implementation details:}
For training, we utilize the Adam optimizer and Cosine Annealing scheduler with an initial learning rate ranging between $1e-5$ and $1e-3$ and set the batch size to $32$ while using L1 loss.
The patch sizes of MTPNet are selected from $\{4, 6, 8, 12, 24, 32, 48, 96\}$ via grid search. 
The look-back window sizes of MTPNet are selected from $\{96, 192, 336, 720\}$ through grid search, except for the ILI dataset, where it was set to $104$.
The transformers of MTPNet consist of 2 encoder layers and 1 decoder layer. The multi-head attention number is set to 4.
The MTPNet is implemented with PyTorch and runs with an NVIDIA GeForce RTX 3090 GPU and an NVIDIA RTX A6000 GPU.
The main results are averaged across six runs with distinct seeds: $1, 2022, 2023, 2024, 2025$, and $2026$. Additionally, ablation studies were conducted using seed $1$.

\subsubsection{Evaluation and baselines:} 
We selected seven state-of-the-art (SOTA) baseline methods as follows:
\begin{itemize}
    \item Transformer-based methods: Informer~\cite{haoyietal-informer-2021}, Pyraformer~\cite{liu2022pyraformer}, FEDformer~\cite{zhou2022fedformer},  Crossformer~\cite{zhang2023crossformer}, PatchTST~\cite{Yuqietal-2023-PatchTST}.
    \item Linear methods: DLinear~\cite{Zeng2022Dlinear}.
    \item CNN methods: MICN~\cite{micnICLR2023}.
\end{itemize}

We obtained the results of DLinear, Pyraformer, Fedformer, and Informer from ~\cite{Zeng2022Dlinear}. The results for PatchTST and Crossformer were obtained from the original paper. In cases where results are not available, we conducted experiments using the optimal hyperparameters as presented in the original papers. In particular, for a fair comparison, we did not fix the input length to 96, as recent studies~\cite{Yuqietal-2023-PatchTST, Zeng2022Dlinear, zhang2023crossformer} have highlighted the significance of optimal input lengths for method performance. 
we utilized Mean Squared Error (MSE) and Mean Absolute Error (MAE) metrics as quantitative measures of forecasting accuracy
Consequently, the results of the baseline methods presented in the experiments represent their respective best performances.

\subsection{Main Results}
Table~\ref{tab:mainres} shows the main experimental results of all methods for nine datasets on MSE and MAE, where the best and second-best results for each case (dataset, horizon, and metric) are highlighted in bold and underlined, respectively.
The MTPNet outperforms SOTA baseline methods, achieving 45 best results and 19 second-best results out of 72 cases.
MTPNet achieves a modest enhancement in accuracy when compared with the best existing method, PatchTST. 
In contrast to DLinear, which raised questions about the effectiveness of transformers in MTS forecasting, MTPNet demonstrates a reduction of 7.04\% in MSE and 8.56\% in MAE.
When compared to the multi-scale CNN-based approach of MICN, MTPNet demonstrates a significant average enhancement of 19.53\% in MSE and 16.72\% in MAE.
Furthermore, when evaluated against the transformer-based multi-scale method Crossformer, MTPNet exhibits a remarkable reduction in MSE and MAE by averages of 39.84\% and 30.32\%, respectively.

\begin{table*}[!htb]
\caption{Quantitative evaluation (MSE/MAE) of state-of-the-art multivariate time series forecasting methods on nine datasets. The forecasting horizons include $24, 36, 48, 96$ for the ILI dataset, and $96, 192, 336, 720$ for the others. Bold results indicate the best performance while underlined results represent the second-best performance.}
\label{tab:mainres}
\centering
\fontsize{7pt}{7pt}\selectfont
\centering
\begin{tabular}{c|c|c|c|c|c|c|c|c|c}
\toprule[1.0pt]
\multicolumn{2}{c|}{Methods}         & {MTPNet} & {PatchTST/64} & {DLinear} & {Crossformer} & {MICN} & {Pyraformer} & {Fedformer} & {Informer} \\
\midrule[0.5pt]
\multicolumn{2}{c|}{Metric}          & MSE~~MAE               & MSE~~MAE                    & MSE~~MAE          & MSE~~MAE          & MSE~~MAE    & MSE~~MAE            & MSE~~MAE             & MSE~~MAE    \\
\midrule[1.0pt]
\multirow{4}{*}{\rotatebox{90}{ETTh$_1$}}	 
      & 	96	 & 	\textbf{0.364~~0.385}  	 &  \underline{0.370}~~\underline{0.400}  	 & 	0.375~~\underline{0.399}	 & 	0.431~~0.458	 & 	0.421~~0.431	 & 	0.664~~0.612	 & 	0.376~~0.419	 & 	0.865~~0.713 \\
      & 	192	 &  \textbf{0.404~~0.410}	 & 	0.413~~0.429	 & 	\underline{0.405}~~\underline{0.416}	 & 	0.420~~0.448	 & 	0.474~~0.487	 & 	0.790~~0.681	 & 	0.420~~0.448	 & 	1.008~~0.792 \\
      & 	336	 & 	\underline{0.431}~~\textbf{0.432}	 &  \textbf{0.422}~~\underline{0.440}	 & 	0.439~~0.443	 & 	0.440~~0.461	 & 	0.569~~0.551	 & 	0.891~~0.738	 & 	0.459~~0.465	 & 	1.107~~0.809 \\
      & 	720	 & 	\underline{0.453}~~\textbf{0.463}	 & 	\textbf{0.447}~~\underline{0.468}	 & 	0.472~~0.490	 & 	0.519~~0.524	 & 	0.770~~0.672	 & 	0.963~~0.782	 & 	0.506~~0.507	 & 	1.181~~0.865 \\
\midrule[0.5pt]
\multirow{4}{*}{\rotatebox{90}{ETTh$_2$}}	 
	 & 	96	 & 	\underline{0.278}~~\textbf{0.335}	 & 	\textbf{0.274}~~\underline{0.337}	 & 	0.289~~0.353	 & 	1.177~~0.757	 & 	0.299~~0.364	 & 	0.645~~0.597	 & 	0.358~~0.397	 &  3.755~~1.525 \\
	 & 	168	 & 	\textbf{0.340~~0.376}	 & 	\underline{0.341}~~\underline{0.382}	 & 	0.383~~0.418	 & 	1.206~~0.796	 & 	0.441~~0.454	 & 	0.788~~0.683	 & 	0.429~~0.439	 &  5.602~~1.931 \\
	 & 	336	 & 	\underline{0.365}~~\underline{0.403}	 & 	\textbf{0.329~~0.384}	 & 	0.448~~0.465	 & 	1.452~~0.883	 & 	0.654~~0.567	 & 	0.907~~0.747	 & 	0.496~~0.487	 &  4.721~~1.835 \\
	 & 	720	 & 	\underline{0.400}~~\underline{0.435}	 & 	\textbf{0.379~~0.422}	 & 	0.605~~0.551	 & 	2.040~~1.121	 & 	0.956~~0.716	 & 	0.963~~0.783	 & 	0.463~~0.474	 &  3.647~~1.625 \\
\midrule[0.5pt] 
\multirow{4}{*}{\rotatebox{90}{ETTm$_1$}}	 
	 & 	96	 & 	\textbf{0.291~~0.332}	 & 	\underline{0.293}~~0.346	 & 	0.299~~\underline{0.343}	 & 	0.320~~0.373	 & 	0.316~~0.362	 & 	0.543~~0.510	 & 	0.379~~0.419	 & 	0.672~~0.571  \\
	 & 	192	 & 	\textbf{0.332~~0.355}	 & 	\underline{0.333}~~0.370	 & 	0.335~~\underline{0.365}	 & 	0.400~~0.432	 & 	0.363~~0.390	 & 	0.557~~0.537	 & 	0.426~~0.441	 & 	0.795~~0.669  \\
	 & 	336	 & 	\textbf{0.367~~0.376}	 & 	\underline{0.369}~~0.392	 & 	0.369~~\underline{0.386}	 & 	0.408~~0.428	 & 	0.408~~0.426	 & 	0.754~~0.655	 & 	0.445~~0.459	 & 	1.212~~0.871  \\
	 & 	720	 & 	0.425~~\textbf{0.410}	 & 	\textbf{0.416}~~\underline{0.420}	 & 	\underline{0.425}~~0.421	 & 	0.582~~0.537	 & 	0.481~~0.476	 & 	0.908~~0.724	 & 	0.543~~0.490	 & 	1.166~~0.823  \\
\midrule[0.5pt]
\multirow{4}{*}{\rotatebox{90}{ETTm$_2$}}	 
	 & 	96	 & 	\textbf{0.164~~0.248}	 & 	\underline{0.166}~~\underline{0.256}	 & 	0.167~~0.260	 & 	0.444~~0.463	 & 	0.179~~0.275	 & 	0.435~~0.507	 & 	0.203~~0.287	 & 	0.365~~0.453  \\
	 & 	192	 & 	\underline{0.223}~~\textbf{0.291}	 & 	\textbf{0.223}~~\underline{0.296}	 & 	0.224~~0.303	 & 	0.833~~0.657	 & 	0.307~~0.376	 & 	0.730~~0.673	 & 	0.269~~0.328	 & 	0.533~~0.563  \\
	 & 	336	 & 	\textbf{0.273~~0.325}	 & 	\underline{0.274}~~\underline{0.329}	 & 	0.281~~0.342	 & 	0.766~~0.620	 & 	0.325~~0.388	 & 	1.201~~0.845	 & 	0.325~~0.366	 & 	1.363~~0.887  \\
	 & 	720	 & 	\textbf{0.356~~0.380}	 & 	\underline{0.362}~~\underline{0.385}	 & 	0.397~~0.421	 & 	0.959~~0.752	 & 	0.502~~0.490	 & 	3.625~~1.451	 & 	0.421~~0.415	 & 	3.379~~1.338  \\
\midrule[0.5pt]
\multirow{4}{*}{\rotatebox{90}{Weather}}	 
	 & 	96	 & 	\textbf{0.146~~0.189}	 & 	\underline{0.149}~~\underline{0.198}	 & 	0.176~~0.237	 & 	0.158~~0.231	 & 	0.161~~0.229	 & 	0.622~~0.556	 & 	0.217~~0.296	 & 	0.300~~0.384  \\
	 & 	192	 & 	\textbf{0.188~~0.230}	 & 	\underline{0.194}~~\underline{0.241}	 & 	0.220~~0.282	 & 	0.194~~0.262	 & 	0.220~~0.281	 & 	0.739~~0.624	 & 	0.276~~0.336	 & 	0.598~~0.544  \\
	 & 	336	 & 	\textbf{0.238~~0.271}	 & 	\underline{0.245}~~\underline{0.282}	 & 	0.265~~0.319	 & 	0.495~~0.515	 & 	0.278~~0.331	 & 	1.004~~0.753	 & 	0.339~~0.380	 & 	0.578~~0.523  \\
	 & 	720	 & 	\textbf{0.310~~0.322}	 & 	0.314~~\underline{0.334}	 & 	0.323~~0.362	 & 	0.526~~0.542	 & 	\underline{0.311}~~0.356	 & 	1.420~~0.934	 & 	0.403~~0.428	 & 	1.059~~0.741  \\
\midrule[0.5pt]
\multirow{4}{*}{\rotatebox{90}{Traffic}}	 
	 & 	96	 & 	\underline{0.401}~~\textbf{0.234}	 & 	\textbf{0.360}~~\underline{0.249}	 & 	0.410~~0.282	 & 	0.538~~0.300	 & 	0.519~~0.309	 & 	0.867~~0.468	 & 	0.587~~0.366	 & 	0.719~~0.391  \\
	 & 	192	 & 	0.431~~\textbf{0.247}	 & 	\textbf{0.379}~~\underline{0.256}	 & 	\underline{0.423}~~0.287	 & 	0.515~~0.288	 & 	0.537~~0.315	 & 	0.869~~0.467	 & 	0.604~~0.373	 & 	0.696~~0.379  \\
	 & 	336	 & 	0.453~~\textbf{0.259}	 & 	\textbf{0.392}~~\underline{0.264}	 & 	\underline{0.436}~~0.296	 & 	0.530~~0.300	 & 	0.534~~0.313	 & 	0.881~~0.469	 & 	0.621~~0.383	 & 	0.777~~0.420  \\
	 & 	720	 & 	0.491~~\textbf{0.284}	 & 	\textbf{0.432}~~\underline{0.286}	 & 	\underline{0.466}~~0.315	 & 	0.573~~0.313	 & 	0.577~~0.325	 & 	0.896~~0.473	 & 	0.626~~0.382	 & 	0.864~~0.472  \\\midrule[0.5pt]
\multirow{4}{*}{\rotatebox{90}{Electricity}}	 
	 & 	96	 & 	\textbf{0.128}~~\textbf{0.219}	 & 	\underline{0.129}~~\underline{0.222}	 & 	0.140~~0.237	 & 	0.141~~0.240	 & 	0.164~~0.269	 & 	0.386~~0.449	 & 	0.193~~0.308	 & 	0.274~~0.368  \\
	 & 	192	 & 	\textbf{0.146}~~\textbf{0.237}	 & 	\underline{0.147}~~\underline{0.240}	 & 	0.153~~0.249	 & 	0.166~~0.265	 & 	0.177~~0.177	 & 	0.378~~0.443	 & 	0.201~~0.315	 & 	0.296~~0.386  \\
	 & 	336	 & 	\underline{0.164}~~\textbf{0.256}	 & 	\textbf{0.163}~~\underline{0.259}	 & 	0.169~~0.267	 & 	0.323~~0.369	 & 	0.193~~0.304	 & 	0.376~~0.443	 & 	0.214~~0.329	 & 	0.300~~0.394  \\
	 & 	720	 & 	0.203~~\underline{0.293}	 & 	\textbf{0.197}~~\textbf{0.290}	 & 	\underline{0.203}~~0.301	 & 	0.404~~0.423	 & 	0.212~~0.321	 & 	0.376~~0.445	 & 	0.246~~0.355	 & 	0.373~~0.439  \\\midrule[0.5pt]
\multirow{4}{*}{\rotatebox{90}{Exchange}}	 
	 & 	96	 & 	0.091~~0.215	 & 	\underline{0.896}~~\underline{0.209}	 & 	\textbf{0.081~~0.203}	 & 	0.323~~0.425	 & 	0.102~~0.235	 & 	1.748~~1.105	 & 	0.148~~0.278	 & 	0.847~~0.752  \\
	 & 	192	 & 	0.175~~\underline{0.301}	 & 	0.187~~0.308	 & 	\textbf{0.157~~0.293}	 & 	0.448~~0.506	 & 	\underline{0.172}~~0.316	 & 	1.874~~1.151	 & 	0.271~~0.380	 & 	1.204~~0.895  \\
	 & 	336	 & 	\underline{0.280}~~\textbf{0.389}	 & 	0.349~~0.432	 & 	0.305~~0.414	 & 	0.840~~0.718	 & 	\textbf{0.272}~~\underline{0.407}	 & 	1.943~~1.172	 & 	0.460~~0.500	 & 	1.672~~1.036  \\
	 & 	720	 & 	\textbf{0.613}~~\underline{0.606}	 & 	0.900~~0.715	 & 	\underline{0.643}~~\textbf{0.601}	 & 	1.416~~0.959	 & 	0.714~~0.658	 & 	2.085~~1.206	 & 	1.195~~0.841	 & 	2.478~~1.310  \\\midrule[0.5pt]
\multirow{4}{*}{\rotatebox{90}{ILI}}	 
	 & 	24	 & 	\underline{1.602}~~\underline{0.837}	 & 	\textbf{1.319~~0.754}	 & 	2.215~~1.081	 & 	3.041~~1.186	 & 	2.684~~1.112	 & 7.394~~2.012	 & 	3.228~~1.260	 & 	5.764~~1.677  \\
	 & 	36	 & 	\textbf{1.371~~0.761}	 & 	\underline{1.579}~~\underline{0.870}	 & 	1.963~~0.963	 & 	3.406~~1.232	 & 	2.667~~1.068	 & 7.551~~2.031	 & 	2.679~~1.080	 & 	4.755~~1.467  \\
	 & 	48	 & 	\textbf{1.371}~~\underline{0.822}	 & 	\underline{1.553}~~\textbf{0.815}	 & 	2.130~~1.024	 & 	3.459~~1.221	 & 	2.558~~1.052	 & 7.662~~2.057	 & 	2.622~~1.078	 & 	4.763~~1.469  \\
	 & 	60	 & 	\underline{1.696}~~\underline{0.884}	 & 	\textbf{1.470~~0.788}	 & 	2.368~~1.096	 & 	3.640~~1.305	 & 	2.747~~1.110	 & 7.931~~2.100	 & 	2.857~~1.157	 & 	5.264~~1.564  \\
\bottomrule[1.0pt]
\end{tabular}
\end{table*}

\subsection{Ablation Study}
\newcommand{\cycletext}[1]{\raisebox{.5pt}{\textcircled{\raisebox{-.9pt} {#1}}}}

\subsubsection{How important is DI embedding?}
Table~\ref{tab:abla_embed} presents a comparison of DI embedding with spatial embedding and temporal embedding. The mechanisms of spatial and temporal embedding are illustrated in Figure~\ref{fig:value_embedding}.
The DI embedding consistently outperformed the spatial embedding mechanism. For the horizon of 96, the MSE and MAE values of the spatial embedding were 9.09\% and 5.6\% higher, respectively, compared to the DI embedding. The performance gap increased further for the horizon of 720, with the spatial embedding showing 32.11\% higher MSE and 12.95\% higher MAE compared to the DI embedding.
The temporal embedding slightly degrades the MSE and MAE values for ETTh1, ETTm1, and Weather datasets by 3.9\% and 3.8\%, respectively. Notably, the temporal embedding achieved the best performance for the Exchange-Rate dataset. We conjecture that this is because the Exchange-Rate dataset inherently exhibits less temporal dependence.
In conclusion, our findings demonstrate that DI embedding outperforms both spatial and temporal embeddings. Furthermore, breaking the dimensionality of MTS data leads to a degradation in performance.

\begin{table}[!htbp]
\caption{Multivariate time series forecasting results of MTPNet with three embedding mechanisms: dimension invariant embedding, spatial embedding, and temporal embedding.}
\label{tab:abla_embed}
\centering
\fontsize{9pt}{9pt}\selectfont
\centering
\begin{tabular}{c|c|c|c|c}
\toprule[1.0pt]
\multicolumn{2}{c|}{Methods}         & {DI} & {Spatial} & Temporal   \\
\midrule[0.5pt]
\multicolumn{2}{c|}{Metric}          & MSE~~MAE               & MSE~~MAE          & MSE~~MAE        \\
\midrule[1.0pt]
\multirow{2}{*}{ETTh$_1$}	 
      & 	96	 & \textbf{0.365}~~\textbf{0.384} & 0.371~~0.396 & 0.369~~0.391 	 \\
      & 	720	 & \textbf{0.455}~~\textbf{0.464}	& 0.554~~0.506 & 0.494~~0.499	 \\
\midrule[0.5pt]
\multirow{2}{*}{ETTm$_1$}	 
	 & 	96	 & \textbf{0.292}~~\textbf{0.333}	& 0.302~~0.339	& 0.302~~0.342	  \\
	 & 	720	 & \textbf{0.424}~~\textbf{0.409}	& 0.431~~0.410	& 0.428~~0.416	  \\
\midrule[0.5pt]
\multirow{2}{*}{Weather}	 
	 & 	96	 & \textbf{0.145}~~\textbf{0.187} & 0.149~~0.193  &0.155~~0.200	  \\
	 & 	720	 & \textbf{0.312}~~\textbf{0.324} & 0.316~~0.326  &0.320~~0.331	  \\
\midrule[0.5pt]
\multirow{2}{*}{Exchange} 
	 & 	96	 & 0.090~~0.214 & 0.116~~0.244 & \textbf{0.085}~~\textbf{0.206}     \\
	 & 	720	 & 0.581~~0.592	& 1.185~~0.840 & \textbf{0.523}~~\textbf{0.548}    \\
\bottomrule[1.0pt]
\end{tabular}
\end{table}

\begin{table*}[!htbp]
\caption{The ablation study results of MTPNet's pyramid structure.}
\label{tab:abla_pyramid}
\centering
\fontsize{9pt}{9pt}\selectfont
\centering
{
\begin{tabular}{c|c|c|c|c|c|c|c}
\toprule[1.0pt]
\multicolumn{2}{c|}{Methods}         & {MTPNet} & {w/o inter-scale} & {w/o all-scale} & {bottom-up} & {Fine} & {Coarse}  \\
\midrule[0.5pt]
\multicolumn{2}{c|}{Metric}          & MSE~~MAE               & MSE~~MAE                    & MSE~~MAE          & MSE~~MAE          & MSE~~MAE          & MSE~~MAE    \\
\midrule[1.0pt]
\multirow{2}{*}{ETTh$_1$}	 
      & 	96	 & 0.365~~\underline{0.384} & 0.365~~0.385 & \underline{0.364}~~0.386 &	\textbf{0.362~~0.383} &	0.400~~0.423 &	0.375~~0.390	 \\
      & 	720	 & 0.455~~0.464	& \underline{0.450}~~\textbf{0.460} & 0.462~~0.462 &	\textbf{0.449}~~0.462 &	0.515~~0.509 &	0.461~~\underline{0.462}	 \\
\midrule[0.5pt]
\multirow{2}{*}{ETTm$_1$}	 
	 & 	96	 & \underline{0.292~~0.333}	& \textbf{0.291~~0.332}	& 0.292~~0.333	& 0.292~~0.333 & 0.307~~0.346 &	0.301~~0.339	  \\
	 & 	720	 & \underline{0.424}~~0.409	& \textbf{0.423~~0.409}	& 0.428~~0.411	& 0.424~~\underline{0.409} & 0.432~~0.417 & 0.428~~0.414	  \\
\midrule[0.5pt]
\multirow{2}{*}{Weather}	 
	 & 	96	 & \textbf{0.145~~0.187} & 0.148~~\underline{0.187}  &0.152~~0.191	& 0.149~~0.195 & 0.158~~0.203 & \underline{0.148}~~0.190	  \\
	 & 	720	 & 0.312~~\underline{0.324} & \textbf{0.309~~0.321}  &0.315~~0.329	& \underline{0.311}~~0.324 & 0.322~~0.331 & 0.313~~0.326	  \\
\midrule[0.5pt]
\multirow{2}{*}{Exchange} 
	 & 	96	 & \underline{0.090~~0.214} & \textbf{0.087~~0.208} & 0.100~~0.223  &	0.096~~0.220&	0.093~~0.218&	0.093~~0.217    \\
	 & 	720	 & \underline{0.581}~~\textbf{0.592}	& \textbf{0.579}~~0.597 & 0.648~~0.619&	0.601~~\underline{0.596}&	0.635~~0.612&	0.648~~0.617      \\
\bottomrule[1.0pt]
\end{tabular}
}
\end{table*}
 
\subsubsection{How important is multi-scale temporal dependency learning?}

To evaluate the effectiveness of multi-scale temporal dependency learning, we present experimental results of single-scale MTPNet of coarse (large patch size) and fine (small patch size) in Table~\ref{tab:abla_pyramid}. Both modifications performed worse than the multi-scale design. Specifically, MTPNet-Fine exhibited a more substantial performance drop than MTPNet-Coarse, showing the challenge of capturing meaningful temporal dependencies from a small number of time steps due to time series data's naturally low semantic characteristics.
Furthermore, the multi-scale transformer pyramid architecture consistently outperformed individual fixed scales. This observation emphasizes the critical importance of the multi-scale transformer pyramid design in the context of MTS forecasting.

\subsubsection{How important are inter-scale connections?}
We modified the MTPNet by removing the inter-scale connections, resulting in no information flow between multiple levels. Each transformer level now receives input solely from the embedded input sequence patches of fixed scale. The results are presented in Table~\ref{tab:abla_pyramid} as ``w/o inter-scale." Surprisingly, the performance of MTPNet without inter-scale connections was even improved by a trivial amount. We conjecture that this improvement may be due to time series data's naturally low semantic characteristics. As a result, MTS forecasting doesn't require latent representation flow between levels to extract high semantic information.

\subsubsection{How important are inputs for all scales?}
The ``w/o inter-scale" column in Table~\ref{tab:abla_pyramid} presents the results of MTPNet without the inputs for all levels. In this configuration, transformers, except for the first level, only take the latent representation from the previous level as input. The performance of MTPNet without all scale inputs dropped for ETTh1 and ETTm1 datasets when the forecasting horizon was $720$ and for both horizons of Weather and Exchange-Rate datasets. 
The latent representations from the lower level are grounded in a different temporal scale. In contrast, the direct MTS data input integrates a DI embedding component, patching input at the specific scale of the current level. As the forecasting horizon increases, the task grows more challenging, emphasizing the greater significance of direct input.
From the experiments, we conclude that direct MTS data input is critical for more challenging forecasting scenarios.

\subsubsection{Top-down vs. Bottom-up.}
The ``bottom-up" column in Table~\ref{tab:abla_pyramid} shows the results of MTPNet with a modification where the top-down latent representation flow in decoders is replaced with a bottom-up approach. This reversal in information flow means that each transformer decoder's input is a fusion of the input sequence and the latent representation of the lower layer (finer scale). 
This modification only degrades the performance of the Exchange-Rate dataset. It is worth noting that MTPNet generates predictions by utilizing a $\operatorname{Conv}$ layer to project from the latent representations of all $K$ levels. Thus, the information from all temporal scales is utilized when generating predictions. This finding indicates that both information flow from coarse to fine and from fine to coarse can enhance forecasting accuracy.

\subsubsection{How does the look-back window length affect the performance?}
\begin{figure}[!htbp]
    \centering
    \includegraphics[width=1\linewidth]{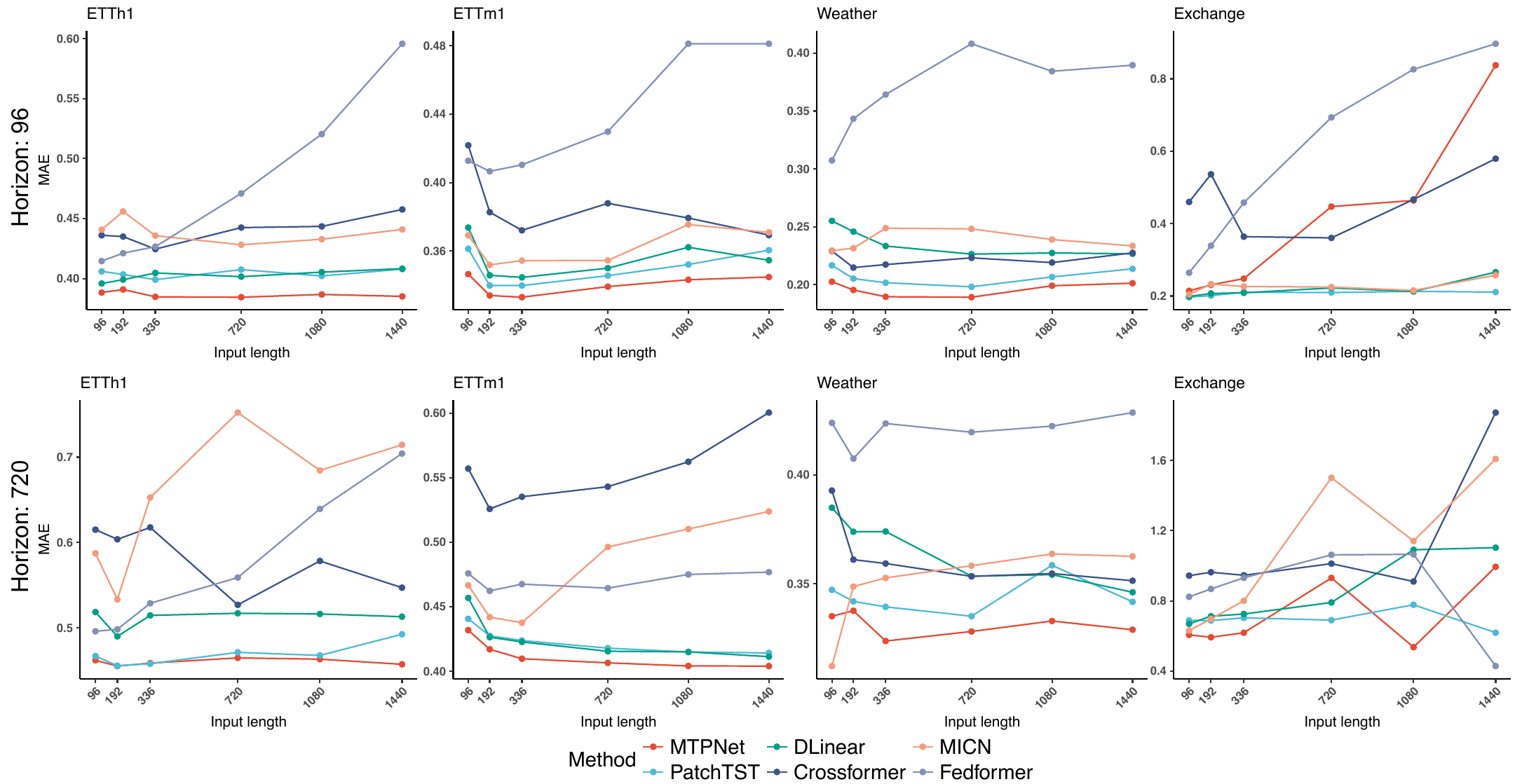}
    \caption{The forecasting results in terms of MAE for different look-back window sizes at horizons 96 and 720.}
    \label{fig:input_len}
\end{figure}

The size of the input sequence plays a crucial role in MTS forecasting as it determines the amount of historical information that can be utilized. Theoretically, an extended input sequence is expected to enhance forecasting accuracy as it encompasses a greater volume of information. However, recent studies~\cite{Yuqietal-2023-PatchTST, Zeng2022Dlinear} have shown that this assumption does not always hold.
Figure~\ref{fig:input_len} illustrates the effect of the input sequence length on the forecasting accuracy. For both horizons, no method consistently benefits from a longer input sequence across all four datasets. In most cases, a method reaches an optimal input sequence length, and its performance degrades when the input sequence becomes longer. Notably, for the Exchange-Rate dataset, a shorter input sequence appears to be optimal for most methods. We attribute this to the inherently low temporal dependency in this dataset. In conclusion, the optimal input sequence length varies depending on the dataset and forecasting horizon.

\section{Conclusion}

This paper introduced a multi-scale transformer-based pyramid network for MTS forecasting. The proposed MTPNet tackles the complexity of modeling temporal dependencies across either a fixed scale or constrained multi-scales. It achieves this by leveraging multiple transformers to capture temporal dependencies at various unconstrained scales. Extensive experimental results demonstrate that MTPNet outperforms existing state-of-the-art methods, particularly those that aim to address the multi-scale temporal dependency issue.

\section{Acknowledgment}

This material is based in part upon work supported by: The National Science Foundation under grant number(s) NSF awards 
    2142428, 
    2142360, 
    OIA-2019609, and 
    OIA-2148788. 
Any opinions, findings, and conclusions or recommendations expressed in this material are those of the author(s) and do not necessarily reflect the views of the National Science Foundation.

\newpage
\bibliographystyle{plain}
\bibliography{mtpnet}

\end{document}